\documentclass[aps,prl,twocolumn,groupedaddress,showpacs,floatfix]{revtex4-1}

\usepackage{dcolumn}
\usepackage{amsmath}
\usepackage{amssymb}
\usepackage{graphicx}
\usepackage{bm}
\usepackage[T1]{fontenc}
\usepackage{color}
\usepackage{url}
\usepackage[bf]{subfigure}
\usepackage{rotating}
\usepackage{soul}
\usepackage{scalefnt}
\usepackage[breaklinks=true,colorlinks=true,linkcolor=blue,urlcolor=blue,citecolor=blue]{hyperref}

\usepackage{color}
\definecolor{redcolor}{rgb}{1.0,0.,0.}

\begin{document}

\title{Classifying informative and imaginative prose using complex networks}
%\shorttitle{Title} %Insert here a short version of the title if it exceeds 70 characters

\author{Henrique F. de Arruda}
%\email{h.f.arruda@gmail.com}
\affiliation{Institute of Mathematics and Computer Science \\
University of S\~ao Paulo, S\~ao Carlos, S\~ao Paulo, Brazil}
\author{Luciano da F. Costa}
%\email{ldfcosta@gmail.com}
\affiliation{S\~ao Carlos Institute of Physics\\
University of S\~ao Paulo, S\~ao Carlos, S\~ao Paulo, Brazil}

\author{Diego R. Amancio}
\email{Corresponding author: diego@icmc.usp.br}
\affiliation{Institute of Mathematics and Computer Science \\
University of S\~ao Paulo, S\~ao Carlos, S\~ao Paulo, Brazil\ }

%\pacs{89.75.Fb}{Preprint submitted to JASIST}%trocar o PACS

\begin{abstract}
Statistical methods have been widely employed in recent years to grasp many language properties. The application of such techniques have allowed an improvement of several linguistic applications, which encompasses machine translation, automatic summarization and document classification. In the latter, many approaches have emphasized the semantical content of texts, as it is the case of bag-of-word language models. This approach has certainly yielded reasonable performance. However, some potential features such as the structural organization of texts have been used only on a few studies.  In this context, we probe how features derived from textual structure analysis can be effectively employed in a classification task. More specifically, we performed a supervised classification aiming at discriminating informative from  imaginative documents. Using a networked model that describes the local topological/dynamical properties of function words, we achieved an accuracy rate of up to 95\%, which is much higher than similar networked approaches. A systematic analysis of feature relevance revealed that symmetry and accessibility measurements are among the most prominent network measurements. Our results suggest that these measurements could be used in related language applications, as they play a complementary role in characterizing texts.
%Because the proposed model is complementary to traditional models, 
\end{abstract}

%In this paper, we have evaluated the ability of network measurements to identify two textual categories, which are related to informative and imaginative documents. We have extended previous models in a twofold manner. First, the local topology of nodes representing specific words was studied. We have thus emphasized particular network regions to characterize the local topology of texts. This approach differs from previous networked representations because  traditional topological analyses consider with equal relevance the topological analysis of all nodes of the network. Another proposed extension is the use of novel network measurements that are able to grasp more relevant information than traditional measurements. Particularly, we have used symmetry measurements that are able to quantify the homogeneity of access to neighbours. The concept of node degree was also extended via introduction of accessibility measurements, which are able to measure the \emph{effective} number of (accessed) neighbours.

\maketitle

\setcounter{secnumdepth}{1}

\section{Introduction}
\label{introduction}

The ever-growing amount of available documents in the Web has propelled the development of statistical natural language processing methods in recent years. Examples of related applications trying to ``understand'' unstructured data include machine translation~\cite{hutchins1992introduction}, text summarization~\cite{amancio2012extractive,yang2008hierarchical,patil2007text,marcu2000theory}, information retrieval~\cite{chen2013text,chen2013text,singhal2001modern} and content analysis~\cite{krippendorff2012content}.  An application of special importance for the organization of electronic data is the classification task, which automatically assigns one or more labels for a word, sentence, paragraph or entire documents~\cite{sebastiani2002machine,aphinyanaphongs2014comprehensive,amancio2012structure,amancio2012identification}. Traditional textual categorization methods usually serve to identify the relevance of texts (i.e. whether it is a spam or not)  or the meaning conveyed by words and expressions~\cite{fetterly2004spam,meyer2004spambayes,navigli2009word,amancio2012unveiling,silva2012word}. Recent classification tasks, however, have emphasized other textual aspects. For example, the categorization of texts according to the their polarity (e.g. positive or negative) has become a relevant task for analyzing e.g. customer reviews~\cite{miao2010fine} or global variations in mood via polarity analysis of twitter messages~\cite{golder2011diurnal}. Note that most of these classification tasks are dependent on text content, since the presence of one or more specific words provides clues about the classes being inferred. While the semantic content is crucial for the success of these applications, the structure of texts might play an important role in classifications problems where the semantics of words is not crucial for the purpose of categorization. This is the case of identifying the style of texts, since documents on the same subject might display different writing styles. In contrast to semantic-based traditional classification tasks, in this paper we probe the relevance of textual structure  to provide useful features for text classification. More specifically, we probe how textual structure depends on two distinct stylistic writing styles: imaginative and informative prose. The structure and organization of texts is studied via networked models, an well known representation of complex systems.

%\bibitem{}
%Krippendorff, Klaus (2004). Content Analysis: An Introduction to Its Methodology (2nd ed.). Thousand Oaks, CA: Sage. p. 413

Networks are discrete models  that basically represent the interrelations between  interacting agents in a complex system. Owing to the simplicity and generality of the model, it has been employed to model a myriad of real and artificial systems~\cite{costa2011analyzing}. Despite being very different in nature, networks modelling distinct systems share several structural patterns~\cite{newman2010introduction}. Of special relevance to this paper, are the networked models of language and texts, which have been useful to unveil universal properties including the scale-free and small-world phenomena~\cite{i2001small}. In practical terms, networked models have been useful to grasp several features of texts, such as quality~\cite{antiqueira2007strong}, complexity~\cite{amancio2012complex} and authenticity~\cite{amancio2013probing}. Particularly, in this study, we used the so-called word adjacency model, which is a approximation of text networks formed by syntactical links~\cite{i2004patterns}. Because the topological analysis of word adjacency networks does not depend on the interpretation of texts, it has been applied with relative success to study the underlying structure of texts,  even when textual content is entirely unknown~\cite{amancio2013probing}. Here we applied such representation to discriminate informative and imaginative prose. We have extended the traditional model in a twofold fashion: (i) we analyzed the local structure of particular nodes (words); and (ii) we considered novel topological/dynamical that are able to unfold the general structure of the network of concepts. As we shall show, both extensions provided competitive classification performance when compared to traditional stylometry methodologies. In special, we have found that symmetries and accessibilities were the most important network measurements. We believe that the proposed extended model could be used to improve the performance of several related problems where the textual structure plays a prominent role in the characterization of documents~\cite{stamatatos2009survey}.

%\bibitem{masucci}
%Network properties of written human language
%A. P. Masucci and G. J. Rodgers
%Phys. Rev. E 74, 026102

This paper is organized as follows. In Section \ref{sec.met}, we present the word adjacency model. Section \ref{sec.patt} describes the pattern recognition methods used to perform the classification. This section also presents a method for measuring feature relevance in a multivariate fashion. The results obtained with  the proposed methodology are described in Section \ref{sec.results}. Finally, Section \ref{sec.conclusion} provides a perspective for further studies and improvements of the model.

\section{Representation and characterization of texts as networks} \label{sec.met}

In this section, we present how a text can be represented as a complex network. We also swiftly describe the main topological network measurements employed to characterize the structure of networks.

\subsection{Modeling texts as complex networks}
\label{complexNetwoks}

A complex network can be represented as a graph, which is defined as a set of nodes and edges~\cite{newman2010introduction}. An usual representation of a network is the adjacency matrix $\boldmath{A} = \{a_{ij}\}$. The elements are defined as:
\begin{equation}
    a_{ij} =
    \begin{cases}
        1 & \text{if there is a link $i \rightarrow j$,} \\
        0 & \text{otherwise.}
    \end{cases}
\end{equation}
In the text mining community, several networked representations of texts have been proposed~\cite{mihalcea2011graph}. If the stylistic properties of texts are relevant for the task being tackled, syntactical relations are employed to establish the links between words~\cite{i2004patterns}. If the application requires the extraction of semantic features, words are connected according to semantic relations, such as those present e.g. in the WordNet~\cite{miller1995wordnet} or in free association graphs~\cite{costa2003what}. In this study, we aim at grasping textual features that are independent of semantic content. For this reason, we used a model that is able to capture stylistic textual features~\cite{cong2014approaching,amancio2012identification}. This model, henceforth referred to as word adjacency model, denotes each distinct word as a node. The edges are established between words appearing in the same context. It has been shown that if one considers the context as the interval of two adjacent words, most of the syntactical relations are recovered~\cite{i2004patterns}. This model has been successfully applied to study many language applications and related systems~\cite{cong2014approaching}.

To create a word adjacency network, usually some pre-processing steps are applied. First, all punctuation marks, line breaks, spaces, numbers and special characters are removed. Particularly, we are mostly interested in the relationship between words conveying semantic information. For this reason,  \emph{stopwords} (or function words) can be optionally removed from the analysis. In the next step, a lemmatization step is performed in order to map words representing
the same concept into the same node.
To assist the lemmatization process, the part-of-speech tag of each word is extracted according to the procedure described in~\cite{greene1971automatic}.
The part-of-speech labelling is required to solve ambiguities, because the same word form might be mapped into distinct lemmas. After the pre-processing step, each remaining distinct word becomes a node and edges are established between adjacent words.

\subsection{Complex network measurements} \label{section.meas}

Currently, there are more than a hundred measurements employed to characterize the topological structure of networks~\cite{costa2007characterization}. Some measurements might depend not only on the structure, but also on a dynamical process (e.g. random walks) occurring on the structure. Below we swiftly describe the measurements used in this study.

\begin{itemize}

 \item \textbf{Number of nodes ($V$):} in a word adjacency network, the number of nodes is the set of different words in the text. In other words, the number of nodes is the vocabulary size of the pre-processed text.

 \item \textbf{Degree ($k$):} the simplest connectivity measurement is the node degree~\cite{boccaletti2006complex},  which corresponds to the total number of edges connected with node $i$. This measurement is defined for directed networks as $k^{(\textrm{in})} = \sum_i a_{ij}$ and $k^{(\textrm{out})} = \sum_j a_{ij}$ for in- and out- degree, respectively. If one considers the undirected and unweighted version of the network, the degree $k_i = \sum_j a_{ji} = \sum_j a_{ij}$ can be understood as the number of distinct bi-grams that a given word appears. If one considers edges weights, then the degree is proportional to the word frequency.

\item \textbf{Neighborhood connectivity ($N$):} this measurement is defined as the number of nodes that can be reached when, starting from the reference node, walks of length $h$ are performed. Note that the traditional degree measurement is recovered when $h=1$.

\item \textbf{Clustering coefficient ($cc$):} given a node $i$, the probability of its neighbors to be connected is called clustering coefficient
($cc_i$)~\citep{watts1998smallworld}. This measurement is defined as
    \begin{equation}
        cc_i =  \frac{3N_\Delta(i)}{N_3(i)},
    \end{equation}
    where $N_\Delta(i)$ is the total number of triangles (i.e. a click comprising three nodes) connected with node $i$, and $N_3(i)$ denotes the number of
connected triples, which is defined as the amount of different connections between $i$ and each pair of nodes. This measurement is traditionally used to quantify the local connectivity of real-world networks~\cite{opsahl2009clustering}. In word adjacency networks, this measurement has been applied to quantify the specificity of words according to the number of distinct contexts in which they appear~\cite{amancio2011comparing}.

\item \textbf{Betweenness centrality ($B$):} to define this measurement, consider all paths connecting any pair of nodes in the network are followed via shortest paths~\cite{Freeman1977Betweenness}. The betweenness of a node $u$ is defined as being proportional to the number of paths that passes through node $u$. More specifically,
    \begin{equation} \label{btweq}
        B_u = \sum_{ij} \frac{\sigma(i,u,j)}{\sigma (i,j)},
    \end{equation}
    where $\sigma(i,u,j)$ is the number of shortest paths between $i$ and $j$ that passes through node $u$ and     $\sigma(i,j)$ is the total amount of shortest paths between $i$ and $j$. According to equation \ref{btweq}, the betweenness centrality can be interpreted as the network flow~\cite{newman2005measure,borgatti2005centrality, freeman1991centrality}, which is a relevant quantity for the analysis of robustness of power-grid networks~\cite{Motter2002Cascade,mirzasoleiman2011cascaded}. When applied to the analysis of text networks, this measurement has been interpreted as being useful to quantify the generality of words in which the word appears~\cite{amancio2015authorship}, which is in part motivated by the use of this measurement in community detection methods~\cite{girvan2002community}. Unlike the clustering coefficient, the betweenness centrality uses the global connectivity information to quantify the specificity/generality of concepts~\cite{amancio2015authorship}.

\item \textbf{Closeness centrality ($C$):} unlike the betwenness centrality, which is based on the \emph{number} of shortest paths, the closeness centrality~\cite{freeman1979centrality} uses the \emph{length} of the shortest paths. If $d_{ij}$ is the shortest distance between nodes $i$ and $j$, the
closeness centrality is calculated as
    %
    %\begin{equation}
        $C_i= {V^{-1}}{\sum_j d_{ij}}$.
    %\end{equation}
    %
Geodesic paths have been reinterpreted in the context of text networks as a measure of word relevance. Actually, a word is deemed relevant if it is very frequent in the text or if it appears related to other very frequent words~\cite{amancio2011comparing}.

\item \textbf{Eccentricity ($E$):} this measurement quantifies the maximum geodesic distance between the reference node and all other nodes~\cite{estrada2011structure}. Therefore, the maximum eccentricity value corresponds to the network diameter. This measurement is calculated for each
node $i$ as
    %
    %\begin{equation}
        $E_i= \max_j( d_{ij})$.
    %\end{equation}

\item \textbf{Eigenvector centrality ($Ec$):} the eigenvector centrality can be understood as an extension of degree centrality~\cite{bonacich1987power}, because the relevance of the reference node relies both on the \emph{number} and \emph{relevance} of neighbors.
%
%In other words, if a vertex has neighbors with high $Ec$ values, its value is increased.
%
Considering the adjacency matrix $\boldmath{A}$, the eigenvector centrality is defined as the eigenvector associated with the leading eigenvalue.
There are many linguistic applications that uses this centrality measurement. It has been applied, for example, in the text summarization task in order to
select the most relevant extracts in texts modelled as graphs~\cite{patil2007text}.

\item \textbf{PageRank ($Pr$):} the PageRank is widely known to be part of the Google's web search~\cite{newman2010introduction,langville2011google}. In texts networks, this measurement has been successfully applied e.g. to disambiguate word senses~\cite{mihalcea2004pagerank}. This measure is based on the eigenvector centrality, and it is defined in matrix terms as
    \begin{equation}
        {Pr} = \alpha {A}{D}^{-1} Pr + \beta {1},
    \end{equation}
    where $\alpha$ and $\beta$ are positive constants (conventionally $\beta=1$), $\boldmath{1}$ is a vector $(1,1,1,\ldots)^T$ and
    $\boldmath{D}$ is a diagonal matrix represented as
    \begin{equation}
    {D}_{ij} =
    \begin{cases}
        \max\{k_i^{(\textrm{out})},1\} & \text{if $i \neq j$,} \\
        0 & \text{otherwise.}
    \end{cases}
\end{equation} 
In contrast with eigenvector centrality, PageRank considers a weighted sum of neighbors importance reflecting the neighbors degree. In this way, the relevance associated to a node is proportionally transferred to its neighbors.

\item \textbf{Accessibility ($A^{(h)}$):} the accessibility  is an extension of the concept of neighborhood connectivity because it measures the effective number of nodes reached at the $h$-th concentric level~\cite{travenccolo2008accessibility}. The effective number of nodes accessed after $h$ steps is computed considering the distribution of probabilities of access via self-avoiding random walks. Mathematically, it is defined using the Shannon entropy~\cite{shannon2001mathematical} of the probabilities of access at the $h$-th concentric level:
    \begin{equation} \label{acessa}
        A^{(h)}_i = \exp \left(- \sum_j p^{(h)}_{ij} \log p^{(h)}_{ij} \right),
    \end{equation}
    where $p^{(h)}_{ij}$ is the probability of a walker starting from $i$ to reach node $j$ in $h$ steps. In text networks, this measurement has been applied to generate summaries and to identify keywords and styles~\cite{amancio2012extractive}. An example of the computation of accessibility is shown in Fig. \ref{acfig}.
    \begin{figure}
\center
\includegraphics[scale=0.45]{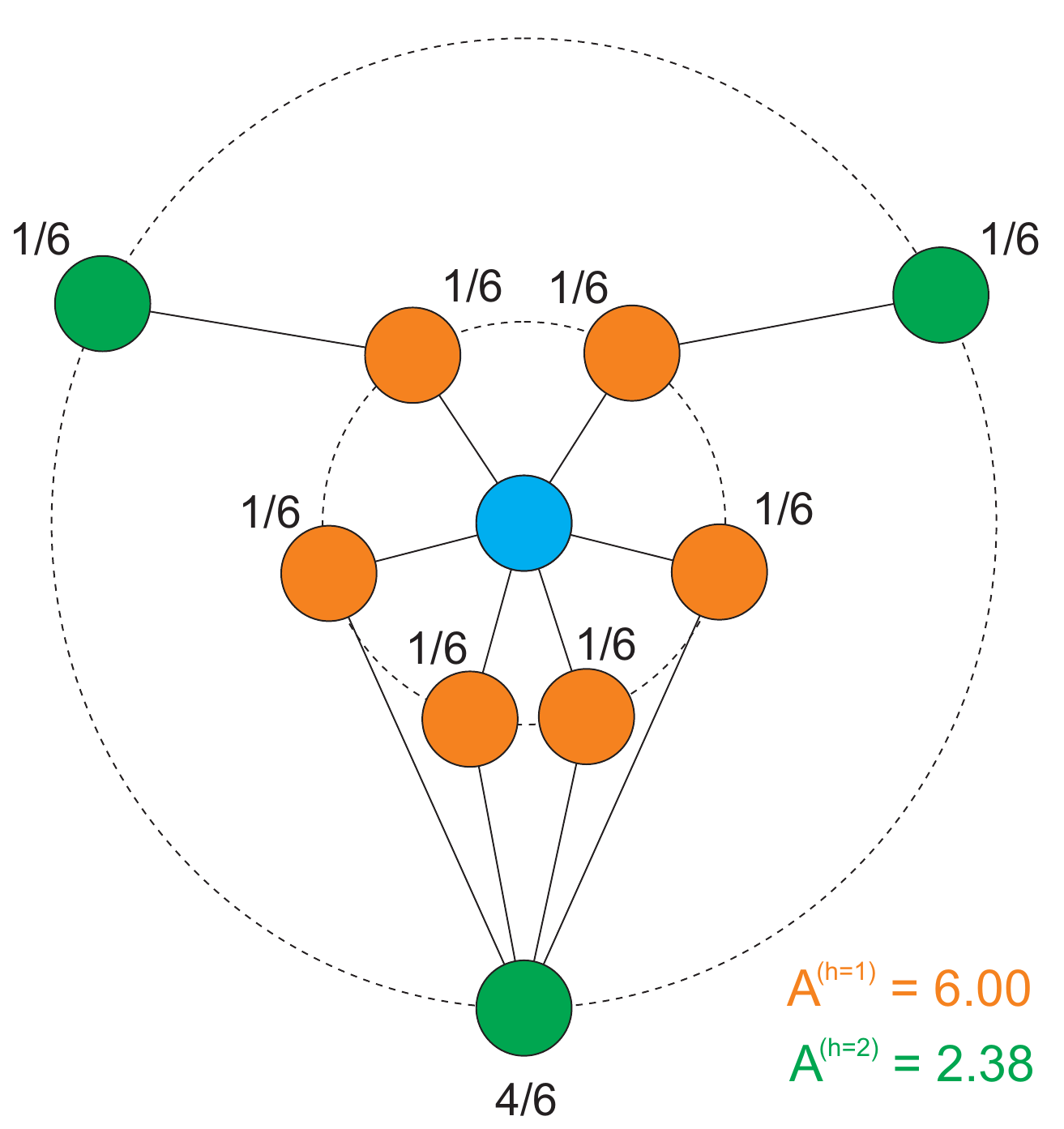}
\caption{\label{acfig}Quantification of accessibility for the local neighbourhood of the blue node. The probabilities near nodes represent the probabilities of access in self-avoiding random walks of length $h=1$ (orange nodes) or $h=2$ (green nodes). In the second hierarchical level, the accessibility is lower than the total number of nodes (green nodes). This occurs because the access to the second level is uneven.}
\label{fig.symmetry}
\end{figure}

\item \textbf{Generalized accessibility ($Ag$):} the generalized accessibility is an extension of the accessibility  that does not rely on a particular length of walk. Instead, the probabilities of transition are computed considering all possible lengths, which is implemented via definition of a modified random walk, the so called accessibility random walk~\cite{arruda2014role}. This measurement is defined as
%
%
%\Red{In order to generalize the accessibility, another measurement was created, the generalized accessibility \cite{arruda2014role}. Through this measurement the parameter $h$ (used in the standard accessibility) is not defined. In other words, $Ag$ is a generalization for every $h$ values. The measurement is defined through a modified random walk, called accessibility random walk, and it is calculated as}
%
%
\begin{equation}
Ag_i = \exp \left(- \sum_j \mathbf{P}_{ij} \log \mathbf{P}_{ij} \right),
\end{equation}
where $\mathbf{P}$ is a quantity that depends on the probability of the transition $i \rightarrow j$ considering walks of variable length. 
The matrix $\mathbf{P}$ representing the probability of transition is calculated as $\mathbf{P} = \mathbf{W}/e$, where
\begin{equation} \label{www}
    \mathbf{W} = \sum_{k=0}^\infty \frac{1}{k!} P^k = e^P.
\end{equation}
Note that, according to equation (\ref{www}), the highest weights are assigned to the nearest nodes. 
The generalized accessibility has been applied e.g. to identify influential spreaders in spatial networks \cite{arruda2014role}.

\item \textbf{Symmetry ($S$):} the symmetry concept is found in many real systems~\cite{finnerty2003origins,longuet1963symmetry}. Symmetric properties can also occur in written texts as a consequence of grammatical or stylistic constraints~\cite{amancio2015concentric}. For this reason, we have quantified this property in word adjacency networks. To model such property, recently, some network measurements have been created~\cite{holme2006detecting,rossi2013characterizing,silva2014concentric,amancio2015concentric}. In this paper, we used the quantities introduced in~\cite{silva2014concentric,amancio2015concentric}, as it allows to capture symmetric patterns in a multi-scale fashion. The definition of symmetry measures rely upon the characterisation of \emph{hierarchic levels}. The hierarchic level $\Gamma_h(i)$ for a given node $i$  is the set comprising all nodes $h$ hops away from $i$. The symmetry measures are based on the accessibility measurement, because the same network dynamics is taken for the analysis. In addition, the symmetry measurement can be seen as a normalization of the accessibility. Thus, using self-avoiding random walks, a node is considered to be symmetric if the access to its neighbors (in a given hierarchic level) is symmetric.  The symmetry (or regularity) of the access is measured in terms of the entropy:
\begin{equation} \label{eq.sim}
  S_h(i) = \frac{\exp{\left\{-\sum\limits_{j\,\in\,\Gamma_{h}(i)} p^{(h)}_{ij} \log  p^{(h)}_{ij} \right\}}}{|\Gamma_{h}(i)| + \sum_{r=0}^{h-1} \eta_r},
\end{equation}
where  $\eta_r$ denotes the total number of dead ends in the $r$-th hierarchical level and $p^{(h)}_{ij}$ is the same quantity used to define the accessibility in equation (\ref{acessa}). There are two variations of the quantity proposed in equation (\ref{eq.sim}). The backbone symmetry ($Sb$), a variation of the the concept of radial symmetry, removes all edges between nodes in the same hierarchical level. The merged symmetry ($Sm$), on the other hand, is based on the concept of angular symmetry, which can be obtained by merging linked nodes in the same hierarchical level. To exemplify both variations of the symmetry concept, we show in Fig. \ref{fig.symmetry} the transformations applied to a hierarchical neighborhood before the computation of equation (\ref{eq.sim}).

% Considering a given node $i$, the measurement is defined in terms of the concentric level $\Gamma_i^{(h)}$, where it is defined as the set of nodes with topological
%distance $h$ from node $i$~\cite{da2008concentric}.
%
%As in accessibility, the symmetry is defined through a dynamics. In this case, it is used a random walk where a walker cannot return to a lower concentric
%level, which is called concentric random walk. The symmetry is compute from Shannon entropy as
%%\begin{table*}
%
%%\end{table*}
%where $P_h(i \rightarrow j)$ is the transition probability between tho nodes $i$ and $j$, given a level $r$, $\eta_r$ is the number of nodes without connections
%with another node from the next concentric level.
%
%In order to consider the edges connecting nodes in the same concentric level, two simplifications were proposed, merged and backbone. Considering the edges connecting nodes in a giving  concentric level, in the first case, the nodes are merged and in the second simplification, the edges are removed, as can be seen in Figure~\ref{fig.symmetry}. In view of those definitions, the symmetry represents two different measurements, the symmetry considering the simplifications merged and backbone, called symmetry merged ($Sm$) and symmetry backbone ($Sb$), respectively. In text networks, the symmetry can be applied in word adjacency Ãnetworks to identify the text authorship.

\begin{figure*}
\center
\includegraphics[scale=0.4]{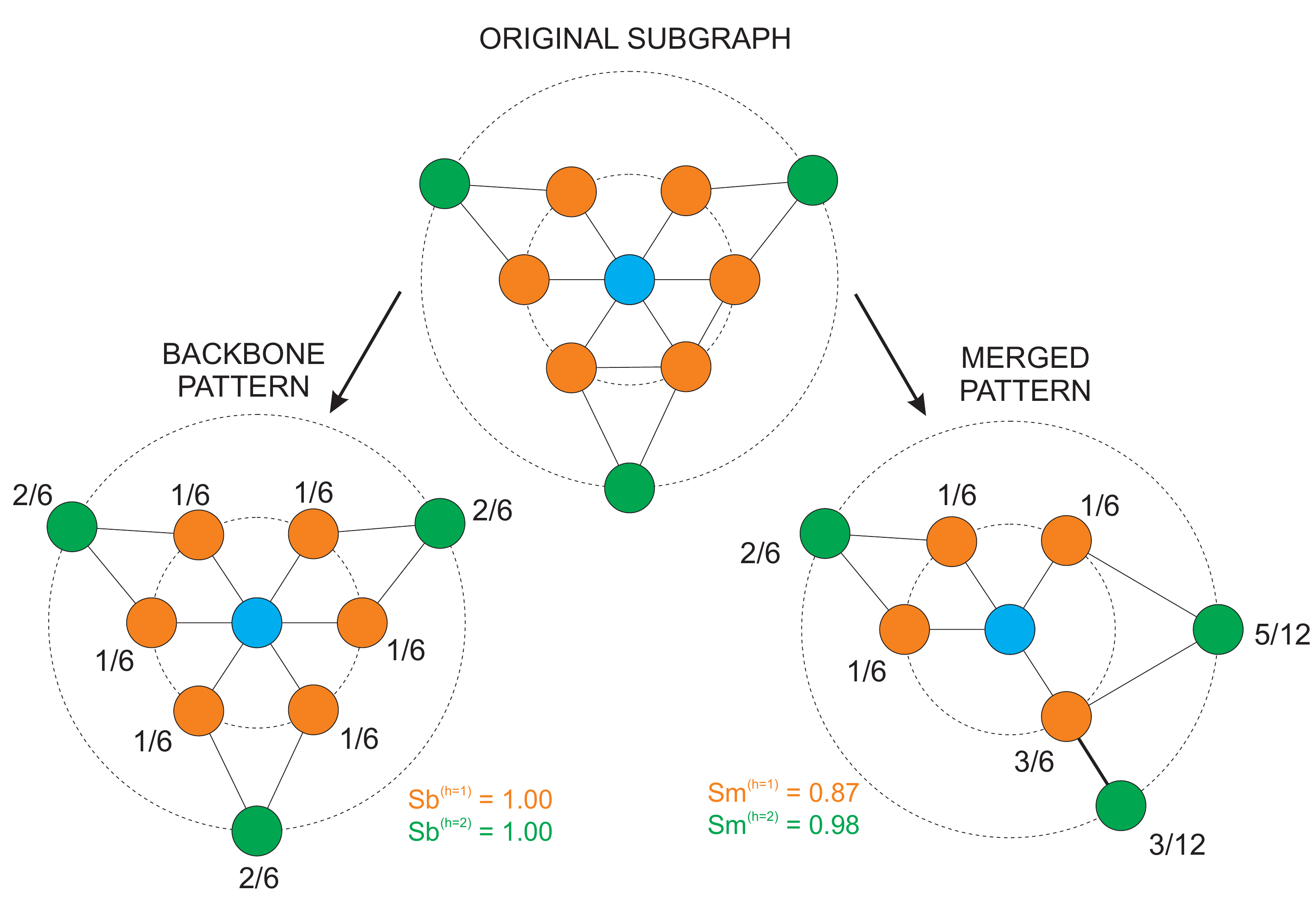}
\caption{Example of merged and backbone patterns for the quantification of local symmetry. The fractions represent the probabilities to reach a given node in a random walk of length $h=1$ (orange) or $h=2$ (green). To create the backbone pattern, edges among nodes in the same hierarchical levels are removed. Differently, merged patterns are computed by merging connected nodes into a single node. }
\label{fig.symmetry}
\end{figure*}

\item \textbf{Modularity ($Q$):} a community structure is defined as a network subgraph with  a large number of intra-links and a few edges connected to the
others nodes of the network. To quantify whether a network is organized in communities, the modularity compares the number of internal links with the expected
value of the same quantity in a equivalent random network~\cite{Newman2006Modularity}. This quantity is computed as
    \begin{equation}
    \label{eq.modularity}
        Q = \frac{1}{2M} \sum_i \sum_j \left( a_{ij} - \frac{k_i k_j}{2M}\right) \delta(c_i,c_j),
    \end{equation}
    where $M = 1/2 \sum_{ij} a_{ij}$ is the total number of edges in the network, $c_i$ and $c_j$ are the communities to which nodes $i$ and $j$ belong, and
    \begin{equation}
    \delta(c_i,c_j) =
        \begin{cases}
            1 & \text{if $c_i = c_j$,} \\
            0 & \text{otherwise.}
        \end{cases}
    \end{equation}
    Usually, word adjacency networks display low values of modularity. A more consistent organization in communities can be found e.g. in semantic networks such as the WordNet~\cite{miller1995wordnet}.

\end{itemize}

\subsection{Characterization of texts with complex networks}

So far we have presented several topological/dynamical measurements of complex networks. The objective here is to use these quantities to characterise styles in texts. Note that several measurements are locally defined, i.e. each node possess a value. There are several possibilities to use these local measurements to characterise the networks. In this paper, we have used the following three distinct methodologies:

\begin{itemize}

\item {\bf Global strategy without stopwords (GS):} in this approach, we sum up the local measurements to characterise the networks. The most natural summarisation procedure is to take the average $\langle X \rangle$, where $\langle X \rangle = V^{-1} \sum_i X_i$ and $X$ is a local measurement. We also used the following quantities: the standard deviation $\sigma(X)$, the median ($\tilde{X}$), the maximum value  ($\max(X)$)  and the minimum value ($\min(X)$).  The only global measurement, the modularity, was also used considered in this strategy.
Following several approaches for grasping textual features with networks, we have removed all stopword before the formation of the networks. These words were disregarded from the analysis because they just serve to connect content words in the word adjacency model.

\item{\bf Local strategy without stopwords (LS)}: in this approach, the value of each measure $X$ for each word becomes a feature. Similarly to the GSW, all stopwords are removed from the analysis. Because features are defined for each word, global networks measurements are not considered in this case.

\item {\bf Local strategy with stopwords (LSS)}: this is the same local approach adopted in the LS method. However, this variation also considers all stopwords in the analysis.

\end{itemize}

\section{Pattern recognition and evaluation} \label{sec.patt}

In this section, we present the methodology for analyzing the relationship between the texts and the categories (informative and imaginative). More
specifically, we describe the pattern recognition methods employed and the methods to compute the quality of the classification and relevance of the proposed
features (see Section \ref{section.meas}).
%
%We show a methodology for pattern recognition, evaluate a classifier performance, and quantify the relevance of features.

\subsection{Pattern recognition methods}
\label{patternRecognition}

To study the relationship between complex network measurements and text style, we used a feature selection algorithm and three different supervised classifiers.
The method used to select the features was the information gain~\cite{mitchell1997machine}, which is a supervised attribute filter known as mutual
information~\cite{Cover2006InformationTheory}. Given the random variables $X$ and $Y$, the mutual information $I(X,Y)$ is computed as
\begin{equation}
    I(X,Y) = \sum_{x \in X}{\sum_{y \in Y}{p(x,y) \log \frac{p(x,y)}{p(x) p(y)}}},
\end{equation}
where $p(x)$ and $p(y)$ are probability functions and $p(x,y)$ is the joint probability.

% \begin{equation}
%     I(X,Y) = H(Y) - H(Y|X) = H(X) - H(X|Y),
% \end{equation}
% %
% where $H(X)$ is the entropy of $X$ and $H(X,Y)$ is the entropy of the joint distribution of values $x \in X$ and $y \in Y$. These quantities are computed as
% %
% \begin{equation}
%     H(X) = -\sum_{x \in X} p(x) \log(p(x)),
% \end{equation}
% %
% \begin{equation}
%     H(X|Y) = H(X,Y) - H(X),
% \end{equation}
% where
% %
% \begin{equation}
%     H(X,Y) = -\sum_{x \in X} \sum_{y \in Y} p(x,y) \log p(x,y).
%     \label{eq:hxy}
% \end{equation}
%
%where $p(x)$ is the probability of the element $x$ in the distribution $X$.  The joint entropy $H(X|Y)$ is calculated as
%where $p(x,y)$ is the
%
The information gain corresponds to the mutual information when $X$ is the values obtained for a given attribute and $Y$ is a vector of corresponding classes.
This technique is used to create a decreasing sorted ranking of relevance. Thus, the most relevant attributes, i.e. the ones with the highest values of
information gain, are selected to perform the classification. An important characteristic of this method is that the attributes are evaluated separately, i.e. the information of a given attribute does not influence the others.

In our experiments, the following  pattern recognition methods were used:
\begin{itemize}
  \item {\bf Nearest neighbors}: the $K$ nearest neighbors classifier ($K$NN) considers the local neighborhood of the test instance~\cite{krzysztof2007data}. Given a test instance, the class chosen is the majority class in the set of the $K$ nearest neighbors in the training dataset. Further details concerning this method can be found in~\cite{bishop2006pattern}.

  \item {\bf Classification and regression tree}: this method represents the patterns found in the dataset as a tree, a data structure storing nested
rules. Even though there are several tree algorithms, we chosen to use the classification and regression tree (CART) method
\cite{breiman1984classification} because it has some advantages as it is relatively simple for interpret and the the predictor variables are not previously
assumed~\cite{lewis2000introduction}. A major advantage of tree-based pattern recognition algorithms the patterns found in the dataset are not hidden from the user, as it happens in artificial neural networks methods~\cite{bishop2006pattern}.

  \item {\bf  Naive Bayes}: the Naive Bayes algorithm is based on the Bayes theorem~\cite{getoor2007introduction}. Assuming feature independence, the correct class $\tilde{c}$ of an instance is given by
      \begin{equation*}
        \tilde{c} = \arg \max_{c_k} \big{[} \log P( c_k )  + \sum_{f_j \in F} \log P( f_j | c_k ) \big{]},
      \end{equation*}
      where $c_k$ is one of the possible classes, $f_j \in F$  is a particular feature. To compute the quantity $P( f_j | c_k )$ we assumed that the likelihood
of the features follows a bell shape~\cite{manning1999foundations}.

\end{itemize}
We have chosen the aforementioned methods because they yield good performance when set with default parameters~\cite{amancio2014systematic}. The evaluation of the performance of the methods when set with default parameters was performed with the ``leave one out'' algorithm \cite{guyon2006feature}. This evaluation procedure consists in selecting one element of the dataset to be used as an test instance, while the remaining instances are used in the training phase. This procedure is then repeated until all instances of the dataset have been chosen as an test instance.

\subsection{Quantifying feature relevance}

To quantify the relevance of features for the classification task, the following method was used. Let  $F=\{f_1,f_2,\ldots\}$ be the set of attributes comprising $\Phi$ distinct attributes. We generate a set $F_c$ comprising all $2^{\Phi}$ combinations of features. For a particular classification method, we compute the accuracy rate obtained for each classifier in $F_c$. The accuracy rate is then employed to sort (in decreasing order) the classifiers in $F_c$. A function is associated for each attribute in $F$:
\begin{equation}
\Omega(f_i,k) = \sum_{j=1}^k \omega(f_i,j)
\end{equation}
where  $\omega(f_i,j) = 1$ if the $j$-th best classifier in $F_c$ used the $i$-th attribute.  If the $i$-th feature was not used in the $j$-th best classifier, then  $\omega(f_i,j) = 0$. Note that the function $\Omega$ quantifies how frequent a given feature is in the best classifiers. If this function has a fast growth for low values of $k$, then it is a relevant feature because it appears in the classifiers with the highest accuracy rates. To quantify how frequent  a given feature $f_i$ is among the best classifiers, the following  index of feature relevance can be defined:
\begin{equation} \label{meu}
	R(f_i) = \sum_{k=1}^{2^{\Phi-1}} \Omega(f_i,k) = \sum_{k=1}^{2^{\Phi-1}} \sum_{j=1}^k \omega(f_i,j).
\end{equation}
Unlike traditional index devised to measure the relevance of attributes, the index defined in eq. (\ref{meu}) takes into account the non-trivial inter-relationship between features~\cite{mann1947test}.

%In order to preselect the features, for each test the we compute the information gain method, but this analysis computes the attributes lonely. Thus, we computed a different methodology that considers the relationship between them. Through all the training database, this method induces the classifiers \cite{amancio2011comparing}. It is executed all possibility of features for a taken classifier, using an algorithm to validate it (in this work we used ``leave one out'' algorithm). Thus, it is possible to evaluate the best attribute set for a taken classifier. Besides, through this methodology it is possible to quantify the feature relevance considering all the features, so showing the relevance of a taken characteristic when it is applied together with others using a Ãmethod called Factor analysis (this method is detailed in \cite{amancio2011comparing}).

\section{Results and discussion}
\label{sec.results}

In this section, we analyze the proposed technique for discriminating informative and imaginative prose. We also compare the proposed technique with other traditional natural language processing methods. In our experiments,  we used the Brown University Standard Corpus of Present-Day American English (a.k.a. Brown Corpus)~\cite{francis1979brown}. Because the set of informative texts comprises several short texts, for this class we have selected only the 126 longest texts. As such, in our experiments, each class is represented by the same number of instances. Each class can also be classified in subclasses. The set of informative texts used in this study comprises 80 {scientific manuscripts}, 30 miscellaneous texts and 16 biographies and related subjects. The set of imaginative documents comprises general fiction, romances, love stories and others. Note that we have not used this fine-grained description in our experiments.

\subsection{Complex network approach}

%To measure the ability of complex network measurements to discriminate informative and imaginative texts,%
%A feature of these texts is that they have tags showing the grammatical function for every token.
%This dataset comprises 374 informative texts and 126 imaginative texts.
%To balance those classes we used 126 samples for each one.
%Each category has some subcategories, so in informative prose we selected 80 texts about learned, 30 miscellaneous texts, and 16 texts about belles lettres, biography, memoirs, etc. In imaginative prose we used all subcategories (general fiction, mystery and detective fiction, science fiction, adventure and western fiction, romance and love story, and humor).

Following the steps in the methodology, we created a word adjacency network for each document in the dataset. The topological measurements were extracted and
the $15$ most relevant features were selected according to the information gain criterion. In the \emph{global strategy}, the following features have been
selected: \\
\ \\
{Vocabulary size}: $V$;\\
Degree connectivity: $\langle k \rangle$; \\
{PageRank}: $\tilde{Pr}$,  $\langle Pr \rangle$, $\sigma(Pr)$ and $\max(Pr)$; \\
{Clustering coefficient}: $\sigma(cc)$ and $\langle cc \rangle$; \\
Closeness centrality:  $\min(C)$, $\sigma(C)$, $\langle C \rangle$ and $\tilde{C}$;  \\
Generalized accessibility: $\langle Ag \rangle$; and \\
Betweenness centrality: $\langle B \rangle$ and $\tilde{B}$. \\
\ \\
In this case, the accuracy rate reached a maximum value of 78\% with the Naive Bayes algorithm. Note that this result is statistically significant, as the
$p$-value associated with this accuracy rate is $p < 1.0 \times 10^{-10}$. The accuracy rate obtained for the other classifiers are shown in the first row of Table \ref{fig.maxresults}.
\begin{table}[t]
\caption{Accuracy rate obtained with the three proposed network approaches. Note that, the most accurate results occur when the }
\label{fig.maxresults}
\begin{tabular}{lccc}
\hline
Complex network approach 		& KNN 	& CART  & Bayes  \\
\hline
Global strategy without stopwords  	& 72\% 	& 78\% 	& 75\% \\
Local strategy without stopwords   	& 92\% 	& 92\% 	& 92\% \\
Local strategy with stopwords     	& 95\% 	& 95\% 	& 95\% \\
\hline
\end{tabular}
\end{table}

\begin{figure}
\center
\includegraphics[scale=0.6]{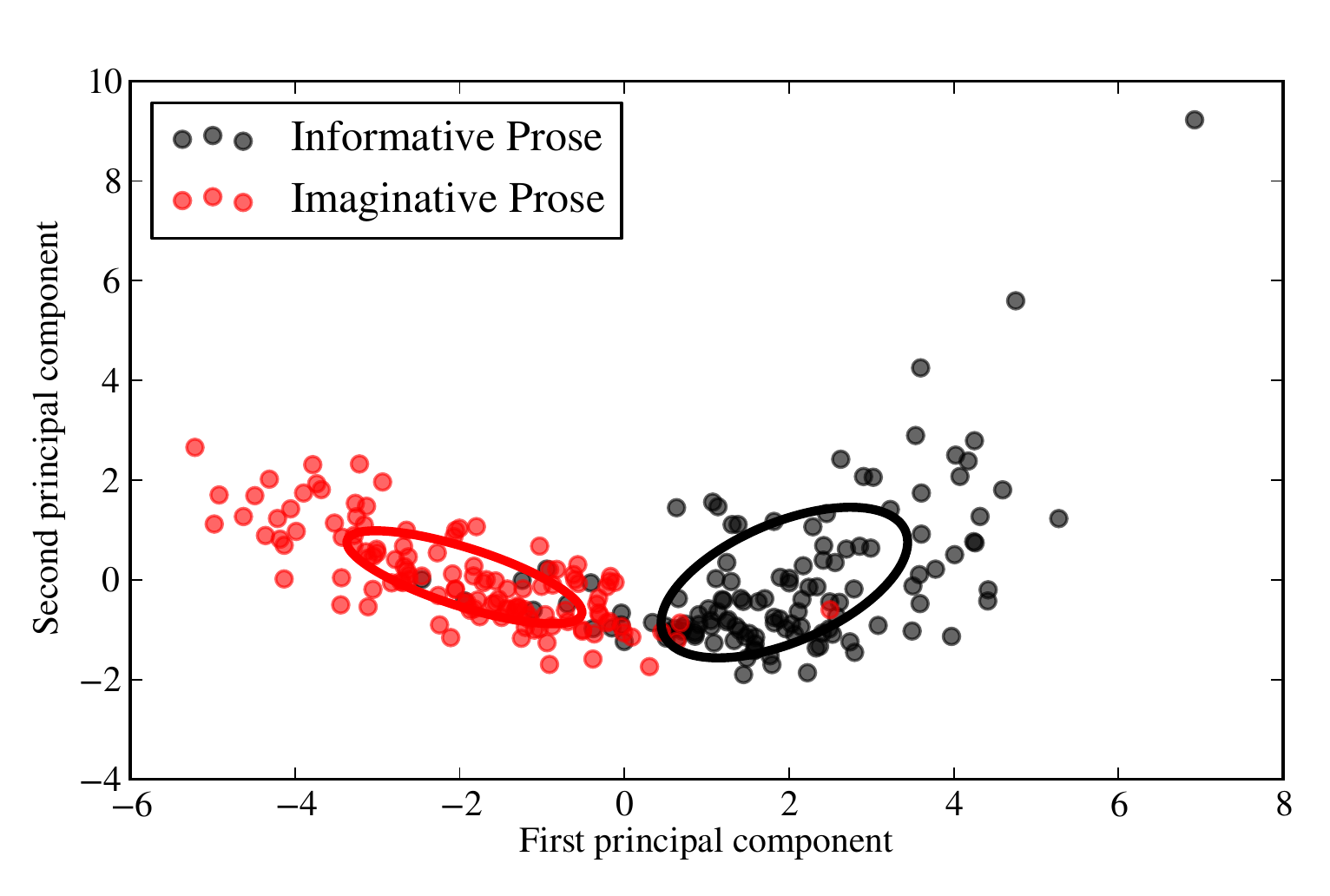}
\caption{Principal component analysis performed using local topological properties in networks formed with stopwords. To create the visualization, only the features with no correlation with the frequency were used.
}
\label{fig.pcaNetworkStop}
\end{figure}

When the \emph{local strategy} (LS) was used to perform the classification, the accuracy rate improved by a large margin: all three classifiers reached an accuracy
rate of 92\%. The largest improvement in performance occurred for the KNN classifier; the accuracy rate went from 72\% to 92\%. The features employed in this
case were:
\ \\
\ \\
Backbone symmetry: $Sb^{(h)}$, for $h=\{2,3,4\}$;\\
Merged symmetry: $Sm^{(h)}$, for $h=\{2,3,4\}$; and \\
Accessibility: $A^{(h)}$, for $h=\{2,3\}$.
\ \\
\ \\
Note that these local measurements were chosen because they do not correlate with the frequency.
The \emph{local strategy with stopwords} (LSS) displayed an slight better classification performance. In this case, the accuracy rate reached 95\% with KNN, CART, and Naive Bayes  classifiers.  The principal component analysis projection provided in Fig. \ref{fig.pcaNetworkStop} confirms the suitability of this network model for
discriminating informative from imaginative prose.
In this case, the features employed were:
\ \\ \ \\
Backbone symmetry: $Sb^{(h)}$, for $h=\{2,3,4\}$; \\
Merged symmetry: $Sm^{(h)}$,  for $h=\{2,3,4\}$; \\
Acessibility: $A^{(h)}$, for $h=2$; and \\
Generalized accessibility: ($Ag$).
\ \\ \ \\
Note that, in both \emph{local strategies}, the accuracy rates in the
classification are much higher than the ones obtained with the \emph{global strategy}, which suggests that a few words account for the informativeness of the topological approach.
To better understand the factors behind the network ability to discriminate informative from imaginative prose, we evaluated the relative importance of features
employed in the best approach, i.e. the \emph{local strategy with stopwords}. The method employed to quantify the relevance of features is described in the
methodology. According to this method, the most relevant features, in decreasing order of relevance were:
\ \\ \ \\
(i)   Merged symmetry: $Sm^{(h=2)}$(\emph{the}) \\
(ii)  Merged symmetry: $Sm^{(h=3)}$(\emph{by}) \\
(iii) Backbone symmetry: $Sb^{(h=4)}$(\emph{by}) \\
(iv) Merged symmetry: $Sm^{(h=3)}$(\emph{an})\\
(v)  Generalized accessibility: $Ag$(\emph{have}) \\
(vi) Merged symmetry: $Sm^{(h=4)}$(\emph{by})  \\
(vii) Generalized accessibility: $Ag$(\emph{it}) \\
(viii) Generalized accessibility: $Ag$(\emph{by}).
\ \\ \ \\
Note that the most relevant features are those related to the symmetry of specific words. Interestingly, this is consistency with recent
results showing that symmetry measurements tend to be more discriminative than other traditional network measurements~\cite{comin2014framework}.

It is relevant to highlight that most of the network approaches for text classification focus on the global properties of networks. Our results reveal, conversely, that the informativeness of the topological strategy concentrates in a few nodes. Particularly, the informativeness was found to be mostly hidden in the symmetry patterns of specific function words. For this reason, we believe that the \emph{local strategies} (LS and LSS) could be useful not only for the studied task, but also in several related tasks, where the topology of specific words plays a prominent role in characterizing texts.

\subsection{Comparison with traditional methods}

%==================================================================================================================================

\begin{figure}
\center
\includegraphics[scale=0.58]{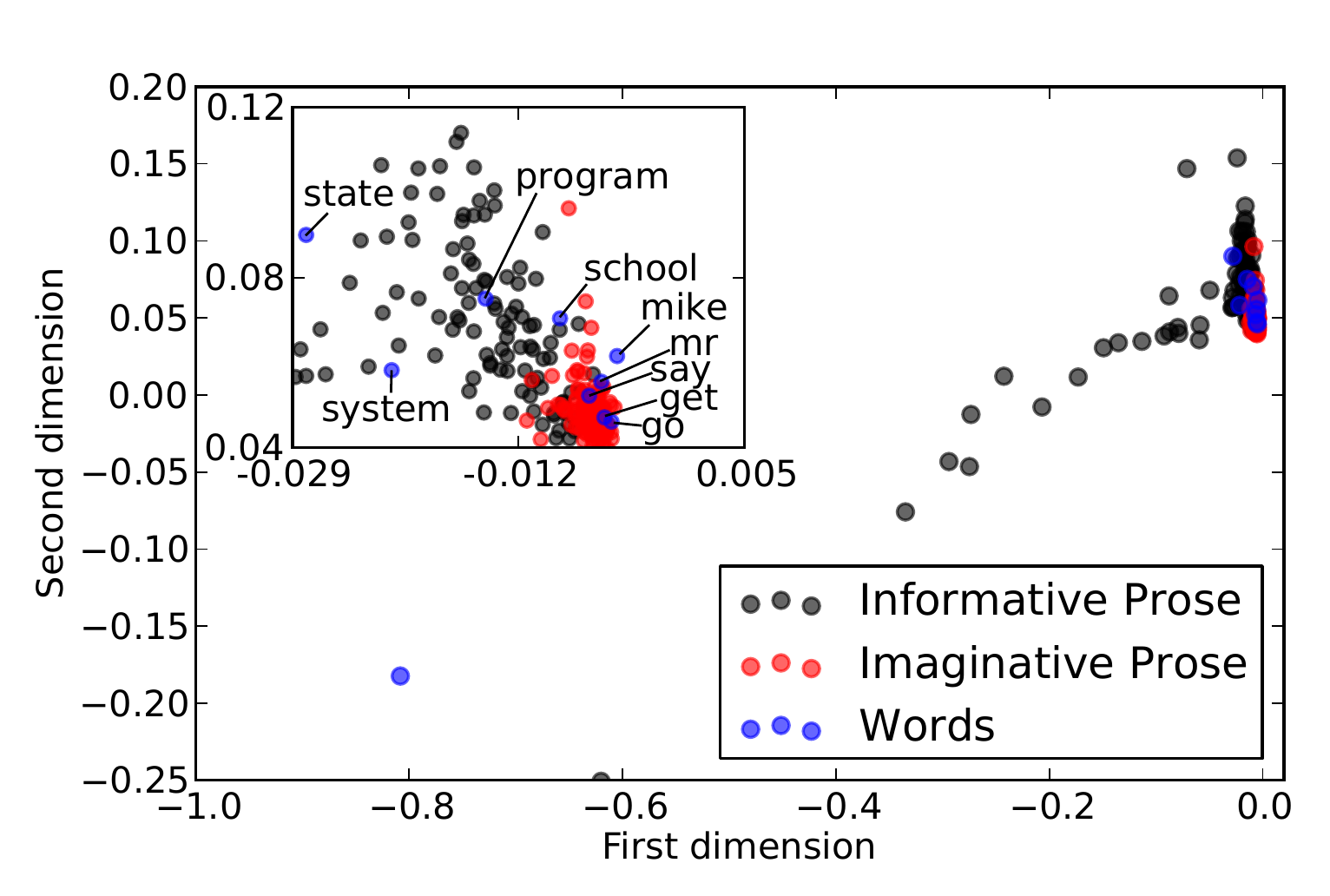}
\caption{Latent semantic analysis performed to distinguish informative from imaginative prose. First ten more frequent words were used as features. Note that the style can be identifies by measuring the proximity to specific words. While \emph{state} and \emph{system} characterize informative documents, \emph{say} and \emph{mr.} characterize imaginative texts.
}
\label{fig.lsa}
\end{figure}

\begin{figure}
\center
\includegraphics[scale=0.58]{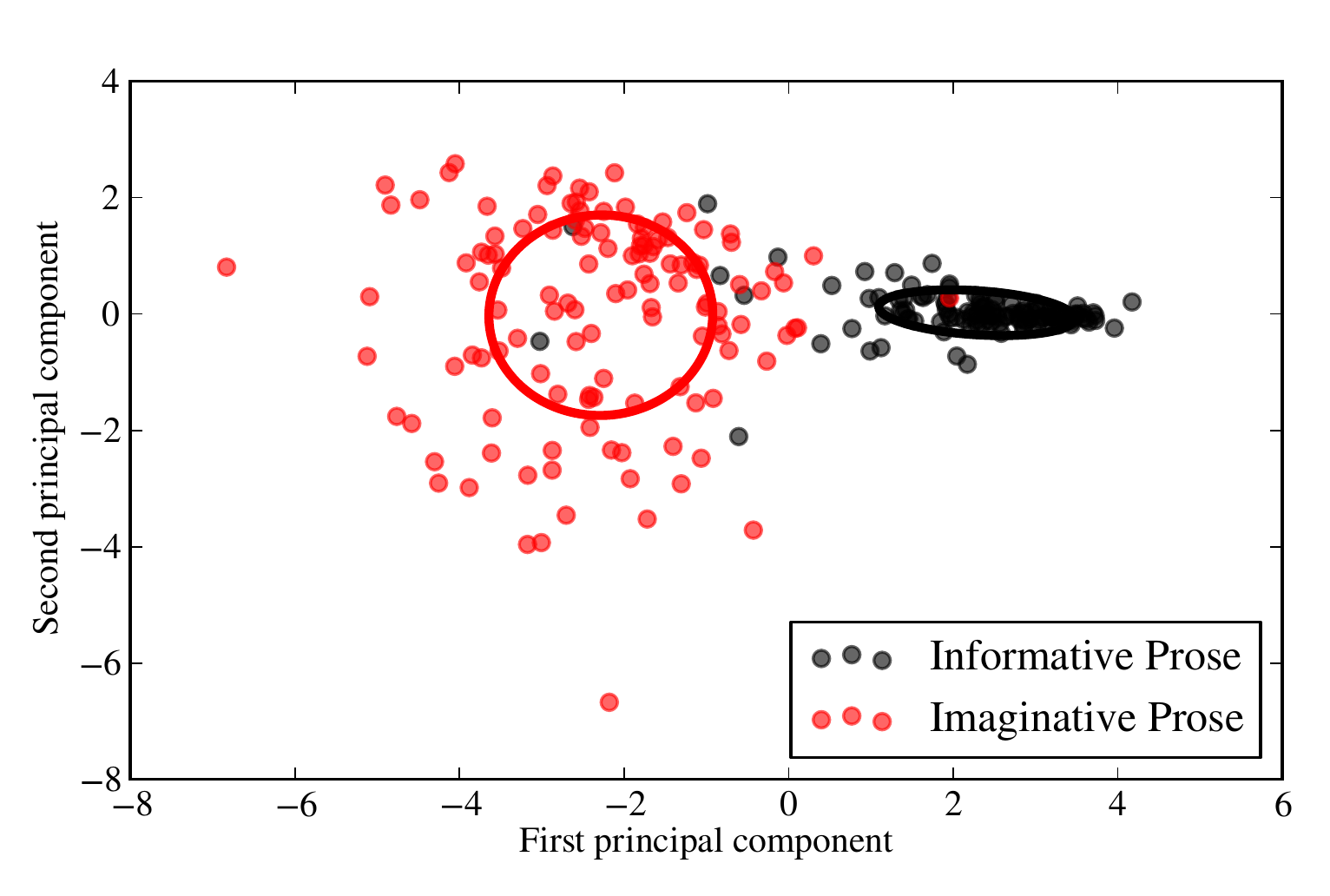}
\caption{Principal component analysis performed to distinguish informative from imaginative prose.  The frequency of the most informative \emph{stopwords} were used as features. Note that the style developed in informative documents is much more regular than the style observed in imaginative texts.}
\label{fig.pcaStop}
\end{figure}

\begin{figure}
\center
\includegraphics[scale=0.58]{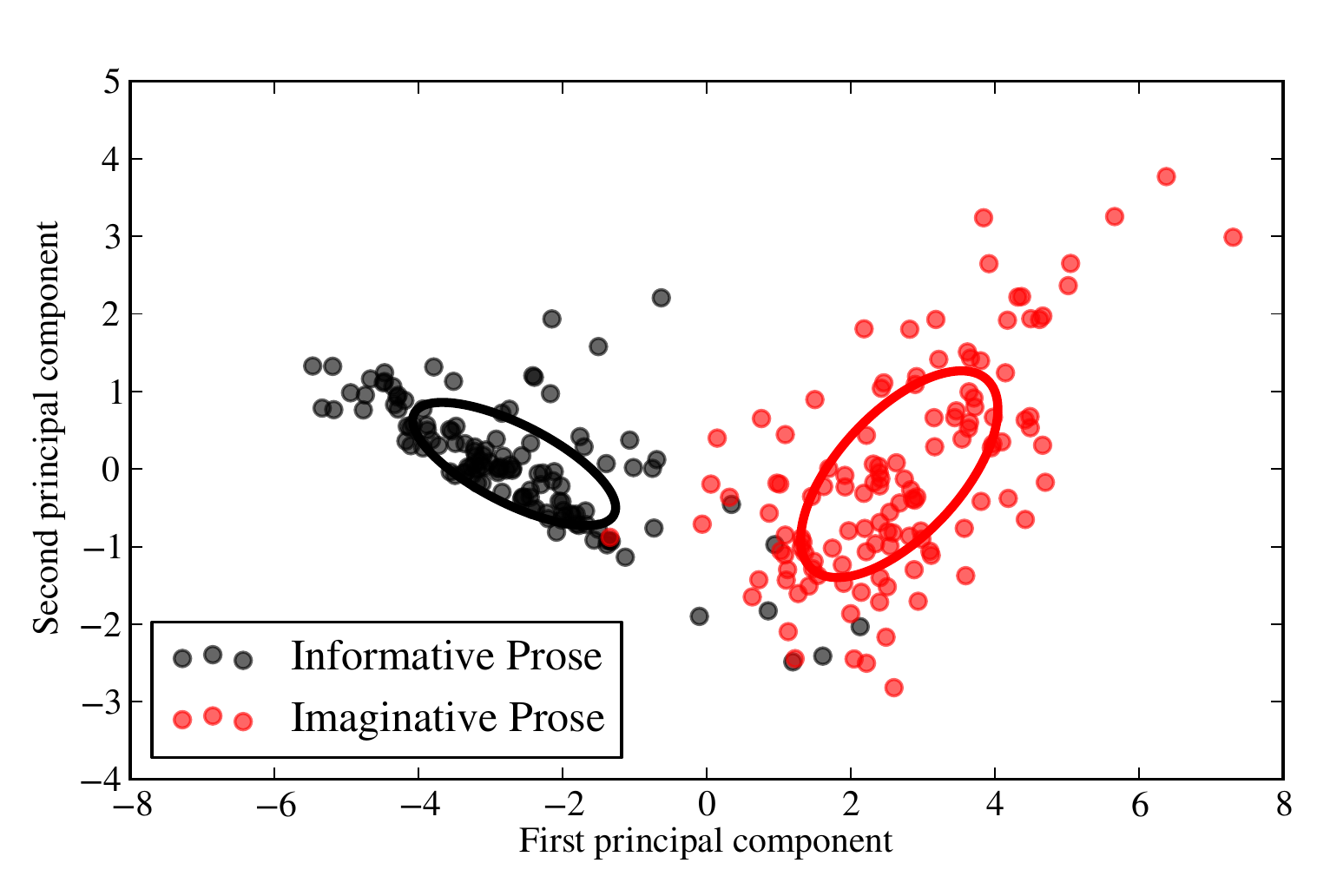}
\caption{Principal component analysis performed to distinguish informative from imaginative prose. The frequency of character bigrams were used as features. As it happened in Fig. \ref{fig.pcaStop}, the variability of styles is much higher in imaginative texts.
}
\label{fig.pcaBigrams}
\end{figure}

To compare the performance of the proposed technique with other traditional techniques, we first analysed if the classes can be discriminated via Latent Semantic Analysis~\cite{dumais2004latent}, which considers as features the frequency of words. The projection obtained with this technique is shown in Fig.~\ref{fig.lsa}. Note that a good discrimination was obtained in this case, mainly because some words are
more common in informative documents (e.g. \emph{state}, \emph{system} and \emph{program}, while others occur more often in imaginative texts (e.g. \emph{say} and \emph{mr.}). A more accurate classification system based on stylistic attributes  can be created if one considers as features the frequency of the most informative \emph{stopwords}. To select the most informative stopwords, we used the information gain criterion. Using the  $K$NN classifier (the best classifier), the performance reached 97\% of accuracy. This high accuracy level can be observed in the principal component analysis provided in Fig.~\ref{fig.pcaStop}.
Another traditional strategy in stylometry consists in counting the frequency of character bigrams~\cite{stamatatos2009survey}. Considering the most informative bigrams, the accuracy rate reached 98\%. A visualisation of the data provided by this set of features is shown in Fig.~\ref{fig.pcaBigrams}.

%Before evaluating our tests, we computed the LSA (Latent Semantic Analysis), which is shown in details in the review \cite{}. This measurement shows the proximity between the meaning of a word and the texts through dimensionality reduction. In order to evaluate the measurement, we only used the words in the text without stopwords and all words were lematized. After this preprocessing, for every text we considered a set of fist words with the size of the smaller text. The result can be seen in figure~\ref{fig.lsa}, where it is possible to observe that the informative prose texts are related to words with general meaning (words like state and system) and imaginative prose texts are related to words specifying actions and people (like say and mr).

%FALANDO DO STOPWORDS
%We used a similar pre-processing as used in the previous analysis, but in this case we kept the stopwords. Through these pre-processed texts, we computed the count of stopwords. In order to consider only the most discriminative attributes, we evaluate the information gain and considered the first $15$ attributes. These results can be seen in the PCA (Principal Component Analysis) \cite{jolliffe1986pca}, in figure~\ref{fig.pcaStop}. Thus, it is possible to note that the measurements of two categories are separated in two regions well defined, but texts categorized as imaginative prose are more disperse than informative %prose.

%FALANDO DE CHARACTERS
%Similar to the stopwords count, we evaluated the frequency of each different bi-gram (considering a bi-gram as two adjacent characters). In order to do a fair comparison, we considered only the first $x$ bi-grams, where $x$ is the amount of bi-grams of the smaller text. We did the same analysis that done in the previous method. Then, we evaluated the information gain and considered the first $15$ attributes. The PCA can be seen in figure~\ref{fig.pcaStop}. Like in stopwords count, seeing the scatter plot of PCA it is possible to observe that the data of imaginative prose is more scattered than informative prose and they are separated in two well defined regions. It shows that the variability amount of bi-grams is grater in imaginative prose.

All in all, the results obtained with traditional classifiers demonstrate that the local topological approach is effective as our best results differs only 3\% from the most efficient traditional system. This result is consistent with similar studies showing that the topology plays a relevant role in characterising complex systems, especially those conveying information~\cite{amancio2015authorship,amancio2012extractive,amancio2012identification,amancio2011comparing}. Because the proposed representation is complementary to the traditional approaches, we advocate that the combination of features of distinct nature (traditional and topological) could lead to the improvement of similar tasks relying on the accurate characterisation of stylistic marks.

\section{Conclusion}
\label{sec.conclusion}

In this paper, we have evaluated the ability of network measurements to identify two textual categories, which are related to informative and imaginative documents. We have extended previous models in a twofold manner. First, the local topology of nodes representing specific words was studied. We have thus emphasized particular network regions to characterize the local topology of texts. This approach differs from previous networked representations because  traditional topological analyses consider with equal relevance the topological analysis of all nodes of the network. Another proposed extension is the use of novel network measurements that are able to grasp more relevant information than traditional measurements. Particularly, we have used symmetry measurements that are able to quantify the homogeneity of access to neighbours. The concept of node degree was also extended via introduction of accessibility measurements, which are able to measure the \emph{effective} number of (accessed) neighbours.

Computational simulations revealed that the proposed extensions are able to improve the efficiency of classification tasks. The best improvement in performance when comparing the traditional model and the proposed method occurred with the $K$NN classifier. An improvement of 23\% was observed, thus confirming the efficiency of the proposed methodology. A systematic analysis of feature relevance revealed that among the most informative attributes are the symmetry and accessibility indexes applied to particular nodes. These results confirm the complementary role played by these measurements in characterizing text networks, since they do not correlate with traditional natural language processing methods. Owing to the generality of the proposed representation and characterization, we believe that it could be extend to a myriad of related applications where the quantification of style is relevant for text categorization. As further works, we intend to combine network methods and traditional statistical methods to improve the performance of the classification. We expect, in this case, that the interwoven combination of methodologies will be able to overcome the limitations of each technique.

\section{acknowledgments}

HFA thanks CAPES for financial support. LdFC is grateful to CNPq (Brazil) (grant no. 307333/2013-2), FAPESP (grant no. 11/50761-2), and NAP-PRP-USP for sponsorship. DRA acknowledges financial support from S\~ao Paulo Research Foundation (FAPESP) (grant no. 14/20830-0).

\bibliography{manuscript}

\end{document}